%% file: main.tex
\documentclass[letterpaper, 10pt, conference]{IEEEtran}

\makeatletter
\long\def\@makecaption#1#2{%
  \vskip\abovecaptionskip
  \sbox\@tempboxa{{\small\bfseries #1.} #2}%
  \ifdim \wd\@tempboxa >\hsize
    {\footnotesize\bfseries #1.} #2\par
  \else
    \global \@minipagefalse
    \hb@xt@\hsize{\hfil\box\@tempboxa\hfil}%
  \fi
  \vskip\belowcaptionskip}
\makeatother

\IEEEoverridecommandlockouts
\usepackage[T1]{fontenc}
\usepackage{inconsolata}
\usepackage{amsmath,amssymb}
\usepackage{bm}
\usepackage{graphicx}
\usepackage{array}
\usepackage{booktabs}
\usepackage{xspace}
\usepackage{xparse}
\usepackage{hyperref}
\usepackage{caption}
\usepackage{subcaption}
\usepackage[table,xcdraw]{xcolor}
\usepackage{pifont}
\usepackage{siunitx}
\usepackage{microtype}
\usepackage{listings}

\definecolor{cNavy}{HTML}{0A2F51}
\definecolor{cBlue}{HTML}{0D427C}
\definecolor{cMist}{HTML}{7C90A2}
\definecolor{cOrange}{HTML}{E8702A}
\definecolor{cFillBlue}{HTML}{E4EDF7}
\definecolor{cGreen}{HTML}{1F9E5A}
\definecolor{cGeo}{HTML}{C0392B}
\definecolor{codebg}{gray}{0.975}

\lstdefinestyle{geodex}{
  basicstyle=\scriptsize\ttfamily\color{cNavy},
  keywordstyle=\color{cBlue}\bfseries,
  commentstyle=\itshape\color{cMist},
  stringstyle=\color{cOrange},
  backgroundcolor=\color{codebg},
  numbers=left, numberstyle=\tiny\color{gray!80}, numbersep=5pt,
  frame=single, framesep=3pt, rulecolor=\color{gray!40},
  breaklines=true, showstringspaces=false,
  columns=fullflexible, keepspaces=true,
  aboveskip=5pt, belowskip=2pt, xleftmargin=14pt,
  captionpos=b,
}
\newcommand{\yes}{\textcolor{cGreen}{\ding{51}}}
\newcommand{\no}{\textcolor{black!22}{\ding{55}}}
\newcolumntype{Y}{>{\centering\arraybackslash}p{3.8em}}

\let\originalleft\left
\let\originalright\right
\renewcommand{\left}{\mathopen{}\mathclose\bgroup\originalleft}
\renewcommand{\right}{\aftergroup\egroup\originalright}

\NewDocumentCommand\Real{}{ \mathbb{R} }
\NewDocumentCommand\T{}{\mathsf{T}}
\NewDocumentCommand\Matrix{m}{ \bm{\mathbf{#1}} }
\NewDocumentCommand\Transpose{m}{ \left.{#1}\right.^\T }
\NewDocumentCommand\LieGroupSO{m}{ \mathrm{SO}(#1) }
\NewDocumentCommand\LieGroupSE{m}{ \mathrm{SE}(#1) }

\newcommand{\geodex}{\mbox{\textit{geodex}}\xspace}
\newcommand{\M}{\mathcal{M}}
\newcommand{\MassMatrix}{\Matrix{M}}
\newcommand{\code}[1]{\texttt{\textcolor{cNavy}{#1}}}

\title{\LARGE \bf \textit{geodex}: A Library for Motion Planning on Riemannian Manifolds}

\author{
Phone Thiha Kyaw$^1{}$, Ben Wei$^1{}$, Sepehr Samavi$^2{}$, Miguel Angel Rogel Garcia$^1{}$ and Jonathan Kelly$^1{}$
\thanks{$^{1}$Authors are with the Space \& Terrestrial Autonomous Robotic Systems (STARS) Laboratory at the University of Toronto Institute for Aerospace Studies (UTIAS), Toronto, Ontario, Canada, M3H~5T6. Email: {\tt\small <first name>.<last name>@robotics.utias.utoronto.ca}}
\thanks{$^{2}$Author is with the University of Toronto Institute for Aerospace Studies (UTIAS) and the Vector Institute for Artificial Intelligence, Toronto, Ontario, Canada. Email: {\tt\footnotesize sepehr@robotics.utias.utoronto.ca}}
}

\begin{document}
\maketitle
\thispagestyle{empty}
\pagestyle{empty}

\begin{abstract}
Planning motions that respect the intrinsic geometry of a robot's configuration space, including its curvature and a configuration-dependent notion of cost, yields shorter, lower-energy, and more natural trajectories than planning under the ambient flat metric.
Existing libraries for optimization on manifolds provide rich geometric primitives but do not plan around obstacles.
While general-purpose motion planning libraries support many state spaces and custom distance functions, they do not yet treat a configuration-dependent Riemannian metric as the geometry that drives distance, interpolation, and geodesics.
We present \geodex, an open-source \mbox{C++20} library with \mbox{Python} bindings.
The library exposes the manifold, its Riemannian metric, the retraction,
and the sampler as independent, interchangeable components through a
single sampling-based motion planning interface.
The same planner runs unchanged on canonical spaces $\Real^n$, $\mathbb{T}^n$, $\mathbb{S}^n$, matrix Lie groups such as $\LieGroupSO{2}$, $\LieGroupSE{2}$, $\LieGroupSO{3}$, and $\LieGroupSE{3}$, products of these spaces, and articulated-robot configuration spaces, each equipped with a user-defined Riemannian metric.
We make \geodex publicly available with documentation, tests, and a reproducible benchmark suite.
\end{abstract}

\input{sections/introduction}
\input{sections/related-work}
\input{sections/design-goals}
\input{sections/architecture}
\input{sections/evaluation}
\input{sections/conclusion}

\section*{Acknowledgements}
The authors would like to thank Sriharsha Ghanta for his early work on the Nav2 plugin and for helpful discussions.

\bibliographystyle{IEEEtran}
\bibliography{references}

\end{document}

%% file: sections/introduction.tex
\section{Introduction}

\looseness=-1
Robot configuration spaces are rarely flat.
The pose of a rigid body lies in the Lie group $\LieGroupSE{2}$ or
$\LieGroupSE{3}$, the configuration of an arm with revolute joints lies
on a product of circles, and task or closed-chain constraints restrict
the configuration to a curved submanifold~\cite{lee2018introduction}.
The cost of moving is also rarely uniform; the kinetic energy of a
manipulator, its distance to the nearest obstacle, and the steering
effort of a wheeled robot all vary with configuration.
Planning with a Euclidean metric ignores both the curvature of the
space and this variation in cost.
A Riemannian metric is an inner product on velocities that varies
smoothly with configuration, so it can represent both.
Shortest paths under a Riemannian metric are \emph{geodesics}, which
generalize straight lines in Euclidean space to curved manifolds.
For a shortest-path planner, the choice of metric therefore determines
the objective.
When the distance function, interpolator and sampler of the planner
are all derived from the same metric, the paths it returns respect the
geometry of the configuration space by construction~\cite{kyaw2026geometry,
jaquier2022riemannian, maric2021riemannian}.

\begin{figure}[t]
\centering
\vspace{-1mm}
\includegraphics[width=0.95\linewidth]{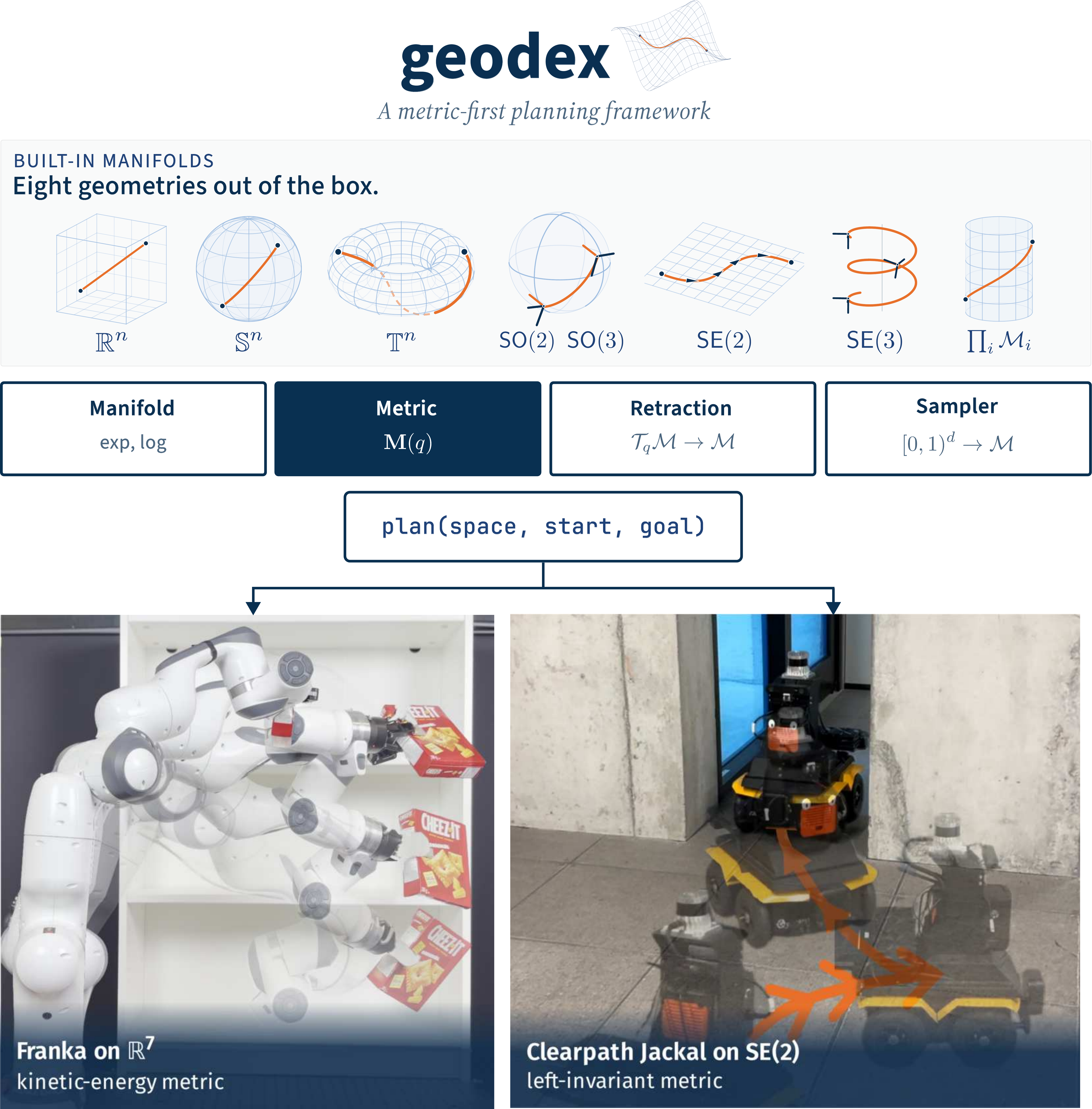}
\caption{
In \geodex, the metric comes first. 
A user builds a space from a manifold and three swappable policies: a metric, a retraction, and a sampler.
We show a Franka manipulator moving a box between shelf compartments under the kinetic-energy metric, and a Clearpath Jackal planning nonholonomic motions on the Lie group $\LieGroupSE{2}$ under a left-invariant metric.
Both tasks use the same planner, and only the \code{space} argument changes.
}
\vspace{-4mm}
\label{fig:infographic}
\end{figure}

\looseness=-1
Existing libraries, however, do not support obstacle-aware planning under a configuration-dependent Riemannian metric.
Geometry libraries~\cite{boumal2014manopt, miolane2020geomstats,pineda2022theseus} let the user define manifolds, metrics, and geodesics, but they solve optimization and estimation problems that do not include obstacles.
Motion planning libraries such as OMPL~\cite{sucan2012open} and its successors~\cite{coleman2014moveit, thomason2024motions} plan around obstacles in many state spaces and accept a custom distance function, but they do not support a configuration-dependent metric.
Planning under a new metric therefore requires changing the planner, not only supplying the metric.

We present \geodex, an open-source library for sampling-based motion planning on Riemannian manifolds.\footnote{
Code is available at \href{https://github.com/utiasSTARS/geodex}{github.com/utiasSTARS/geodex}, with documentation and tutorials at \href{https://geodex.readthedocs.io}{geodex.readthedocs.io}.
}
\geodex builds on established libraries, taking planners from OMPL~\cite{sucan2012open}, dynamics from Pinocchio~\cite{carpentier2019pinocchio}, and fast collision checking from VAMP~\cite{thomason2024motions} to create a single planning interface with the robot manifold, its Riemannian metric, the retraction, and the sampler as independent and interchangeable components. This way, the Riemannian metric defines the geometry that shapes every stage of planning, so a change of metric propagates through the whole planner at once.

Our contributions are threefold. First, we present \geodex, a metric-first planning framework, which provides a single planning interface across many robot manifolds and metrics. Second, we extensively evaluate the capabilities of \geodex on two different robot platforms, a Franka manipulator and a Clearpath Jackal mobile robot (Figure~\ref{fig:infographic}), in both offline and real-world experiments. Our manipulation experiment uses \geodex for a shelving task and we show that swapping the metric alone changes how the Franka arm moves a box between shelf compartments.
Our mobile robot experiments compare \geodex with existing planners for nonholonomic navigation on $\LieGroupSE{2}$. We show that planning on the manifold yields paths that are smoother and that can be traversed more quickly than the ones the existing planners build from discretized motion primitives.
Third, we make \geodex publicly available with documentation, tutorials, and tests, along with the integration of our framework into popular robotics software stacks through MoveIt2 and Nav2 plugins.

%% file: sections/related-work.tex
\begin{table*}[t]
\centering
\caption{
Comparison of widely used motion-planning and geometry libraries.
Planning libraries offer obstacle-aware, sampling-based search but treat geometry as a fixed ambient metric, whereas geometry libraries offer configuration-dependent Riemannian metrics and geodesics but do not plan around obstacles.
\geodex is the only library that provides both, combining collision-aware, asymptotically optimal sampling-based planning with a first-class, configuration-dependent Riemannian metric, informed search, low-discrepancy on-manifold sampling, and both C++ and Python interfaces.}
\label{tab:capabilities}
\footnotesize
\setlength{\tabcolsep}{4pt}
\renewcommand{\arraystretch}{1.125}
\begin{tabular}{l *{10}{Y}}
\toprule
& \multicolumn{3}{c}{\textbf{Planning}} & \multicolumn{3}{c}{\textbf{Geometry}} & \multicolumn{2}{c}{\textbf{Sampling}} & \multicolumn{2}{c}{\textbf{Interface}} \\
\cmidrule(lr){2-4}\cmidrule(lr){5-7}\cmidrule(lr){8-9}\cmidrule(lr){10-11}
&
\shortstack{~\\Obstacles} & \shortstack{Sampling-\\based} & \shortstack{Asympt.\\optimal} &
\shortstack{Riem.\\metric} & \shortstack{Geodesic\\paths} & \shortstack{Multiple\\manifolds} &
\shortstack{~\\Informed} & \shortstack{Low-\\discrep.} &
\shortstack{~\\C++} & \shortstack{~\\Python} \\
\midrule
OMPL~\cite{sucan2012open}                                     & \yes & \yes & \yes & \no  & \no  & \yes & \yes & \no  & \yes & \yes \\
MoveIt~\cite{coleman2014moveit}                               & \yes & \yes & \yes & \no  & \no  & \yes & \yes & \no  & \yes & \yes \\
VAMP~\cite{thomason2024motions}                               & \yes & \yes & \yes  & \no  & \no  & \no  & \yes  & \yes  & \yes & \yes \\
cuRoboV2~\cite{sundaralingam2026curobov2}                     & \yes & \no  & \no  & \no  & \no  & \no  & \no  & \no  & \yes & \yes \\
PyRoki~\cite{kim2025pyroki}                                   & \yes & \no  & \no  & \no  & \no  & \no  & \no  & \no  & \no  & \yes \\
Manopt/Geomstats~\cite{boumal2014manopt,miolane2020geomstats} & \no  & \no  & \no  & \yes & \yes & \yes & \no  & \no  & \no  & \yes \\
RMPflow~\cite{cheng2018rmpflow}                               & \yes & \no  & \no  & \yes & \no  & \no  & \no  & \no  & \no  & \no  \\
GGCS~\cite{cohn2025non}                                       & \yes & \no  & \yes & \no  & \yes & \no  & \no  & \no  & \yes & \yes \\
\midrule
\rowcolor{cFillBlue}
\geodex                                                       & \yes & \yes & \yes & \yes & \yes & \yes & \yes & \yes & \yes & \yes \\
\bottomrule
\end{tabular}
\vspace{-3mm}
\end{table*}

\vspace{-1mm}
\section{Related Work}
\label{sec:related}

\looseness=-1
\emph{Motion planning libraries:} OMPL~\cite{sucan2012open} is the de facto sampling-based motion planning library.
It provides many state spaces, including $\LieGroupSO{2}$, $\LieGroupSO{3}$, $\LieGroupSE{2}$, $\LieGroupSE{3}$, and compound products, each with built-in distance and interpolation subroutines, and it lets a user supply custom distance functions and optimization objectives.
The OMPL constrained-planning framework~\cite{kingston2019exploring} plans on implicitly defined manifolds and approximates geodesics on the constraint surface but measures motion under the ambient Euclidean metric rather than a metric intrinsic to the robot manifold itself.
However, OMPL does not provide a configuration-dependent Riemannian metric that jointly induces the distance, the geodesic interpolation, and admissible heuristics.
\geodex supplies this Riemannian layer as an addition to OMPL's mature planners underneath, enabling the planning algorithms to be applied to richer state spaces.

Other motion planning frameworks that build on top of OMPL, such as MoveIt~\cite{coleman2014moveit}, package OMPL's planners for manipulators but inherit the same state model.
High-performance libraries such as VAMP~\cite{thomason2024motions} accelerate sampling-based planning with SIMD vectorization, while the GPU optimizers cuRobo~\cite{sundaralingam2023curobo} and cuRoboV2~\cite{sundaralingam2026curobov2} generate manipulator trajectories through parallel local optimization.
cuRoboV2 enforces torque limits through inverse dynamics and adds a mechanical-work cost that the authors label as energy, yet that energy is a soft constraint in a single robot's joint space rather than an intrinsic metric, and, unlike \geodex, these systems target articulated manipulators instead of arbitrary manifolds.
PyRoki~\cite{kim2025pyroki} shares our modular philosophy as a JAX toolkit for kinematic optimization, covering inverse kinematics, trajectory optimization, and motion retargeting, but it provides neither sampling-based planning nor support for a configuration-dependent Riemannian metric.
Optimization-based frameworks such as CHOMP~\cite{zucker2013chomp}, STOMP~\cite{kalakrishnan2011stomp}, and Crocoddyl~\cite{mastalli2020crocoddyl} optimize trajectories under smoothness and dynamics costs but are not metric-agnostic planning libraries.

\emph{Geometry libraries:} %
\looseness=-1
Riemannian optimization toolboxes such as Manopt~\cite{boumal2014manopt}, Pymanopt~\cite{townsend2016pymanopt}, and Geomstats~\cite{miolane2020geomstats}, together with differentiable and Lie-group libraries such as Theseus~\cite{pineda2022theseus} and manif~\cite{deray2020manif}, provide manifolds, metrics, retractions, and solvers.
They target optimization and state estimation primarily, and have no notion of obstacles, or collision-free motion planning.

\looseness=-1
\emph{Geometry-aware planning and control:} Reactive Riemannian methods such as RMP~\cite{ratliff2018riemannian} and RMPflow~\cite{cheng2018rmpflow} use metrics to reshape local policies rather than to plan globally.
Non-Euclidean global planners such as GGCS~\cite{cohn2025non} use graphs of geodesically convex sets but are restricted to specific geometries and convex decompositions.
Recent geometry-aware methods that combine sampling-based planning with Riemannian metrics~\cite{kyaw2026geometry, li2024riemannian} are published as algorithms rather than as a reusable, metric-agnostic library.
\geodex packages these capabilities behind a single interface (Table~\ref{tab:capabilities}).
To the best of our knowledge, \geodex is the first reusable library in which a user-defined Riemannian metric is the central object used in collision-aware, sampling-based motion planning.

%% file: sections/design-goals.tex
\section{Design Goals and System Overview}
\label{sec:design-goals}

\looseness=-1
Motion planning algorithms typically rely on several subroutines, such as distance functions or a cost functional, local steering methods, and nearest-neighbor queries to find collision-free paths.
On a Riemannian manifold, these components must all be consistent with the underlying geometry.
In most planning libraries, however, they are implemented separately and assume a fixed metric, most often the ambient Euclidean one, so adapting a planner to a new geometry means reimplementing each component by hand.
The goal of \geodex is to make planning under a configuration-dependent Riemannian metric just as simple to express and just as efficient to run as planning under the Euclidean metric.
The principles below follow from this goal.

\looseness=-1
We treat the Riemannian metric as a first-class citizen that shapes every stage of planning, rather than as a cost term added only at the end.
A single metric induces the distance, steering, and cost computations that planning  algorithms utilize, so a change of metric needs to propagate across all components.
To make this practical, we compose each manifold from a base topology and three interchangeable policies.
The base topology provides the exponential and logarithmic maps, the metric defines the geometry through its inner product and induced distance, the retraction and its inverse move points back and forth between tangent spaces and the manifold, and the sampler determines how configurations are drawn.
Each manifold also carries a measure-preserving map from the unit cube, so a single low-discrepancy sampler produces well-spread points on any manifold under its intrinsic measure.
Because these policies are decoupled, any one can be replaced without modifying the others.
For example, a Euclidean space can be equipped with either a canonical flat metric or a configuration-dependent kinetic-energy metric, and a sphere with either an exact exponential map or a cheaper first-order projection, without changing any other part of the planner.

This decoupling is realized through \mbox{C++20} concepts and template policies that resolve at compile time.
An algorithm is written once against the \code{RiemannianManifold} concept and specializes to each concrete manifold and metric through static dispatch, with no virtual calls and no runtime registration.
The core library is a lightweight, template-based collection of geometry and planning code, while robot dynamics and SIMD collision checking are provided as optional compiled modules.
This design also keeps the library extensible, because a new manifold or metric is simply a small class that satisfies a concept, with no base class to inherit and no factory to register, and it works with every planning algorithm once compiled.
To keep development and prototyping convenient, \geodex supports \mbox{Python} bindings out-of-the-box, so a metric or planning query can be prototyped in a few lines of code while the underlying computation still runs in the compiled \mbox{C++} core.
Finally, rather than reimplementing existing tools, \geodex builds on established libraries. It uses the sampling-based planners from OMPL, obtains robot dynamics from Pinocchio, and relies on VAMP for fast SIMD collision checking, with integration into open-source robotics stacks through Nav2 and MoveIt~2 plugins.
Section~\ref{sec:architecture} describes how these principles are implemented as concepts and policies in detail.

%% file: sections/architecture.tex
\section{Architecture}
\label{sec:architecture}

The design goals of Section~\ref{sec:design-goals} demand a single planner that runs under \emph{any} Riemannian metric at close to the speed of Euclidean planning.
\geodex achieves this by expressing every manifold in terms of one base topology and three interchangeable policies (metric, retraction, and sampler) that a single generic planner consumes through one compile-time interface.
We first describe that interface and the static dispatch that makes it fast (Section~\ref{subsec:concepts}), followed by the sampler policy that spreads low-discrepancy samples across each manifold (Section~\ref{subsec:sampling}), the metrics, admissible heuristics, and planners that define the geometry and search the manifold for shortest paths (Section~\ref{subsec:metrics}), and finally the pre-compiled dynamics, collision, and integration modules that bring metric planning to real robots (Section~\ref{subsec:performance}).

\begin{figure*}[t]
\begin{lstlisting}[language=C++]
// (a) C++: the base topology and the injected metric are template arguments.
using Manipulator = ConfigurationSpace<Euclidean<7>, KineticEnergyMetric<MassFn>>;
static_assert(RiemannianManifold<Manipulator>);   // the compiler verifies the composition, no base class

Manipulator manipulator{Euclidean<7>{}, KineticEnergyMetric<MassFn>{mass}};  // inject the metric M(q)
auto path = plan(manipulator, start, goal).path;                             // same planner, statically dispatched
\end{lstlisting}
\vspace{-4pt}
\begin{lstlisting}[language=Python]
# (b) Python: the same manipulator, built from one metric callback; the geometry runs in the C++ core.
import geodex

def mass(q): ...   # a configuration-dependent metric, q -> M(q) (SPD)

metric      = geodex.KineticEnergyMetric(mass)                        # geometry from a single callback
manipulator = geodex.ConfigurationSpace(geodex.Euclidean(7), metric)  # R^7 manipulator, kinetic-energy geometry
path        = geodex.plan(manipulator, start, goal).path              # one call, runs on the compiled core
\end{lstlisting}
\caption{
Composing a manifold in \geodex.
Both panels build the same articulated manipulator, with a Euclidean topology carrying a kinetic-energy metric $\MassMatrix(q)$, and plan on it.
(a) In \mbox{C++}, a type models the \code{RiemannianManifold} concept when it provides the topology and geometry operations, \code{ConfigurationSpace} injects a metric onto a base topology, and the composition is verified by the compiler.
Every geometric call the planner makes is then statically dispatched and inlined rather than routed through a virtual call.
(b) The same manipulator in Python, where the configuration-dependent metric enters as one callback $q\mapsto\MassMatrix(q)$ and a single \code{plan} call runs the identical planning query while the geometry still executes in the compiled \mbox{C++} core.}
\label{fig:compose}
\vspace{-3mm}
\end{figure*}

\subsection{Concepts and policy-based design}
\label{subsec:concepts}

\looseness=-1
A \emph{concept} in \mbox{C++20} is a named set of requirements that the compiler checks a type against.
In \geodex, a type models the \code{RiemannianManifold} concept when it exposes the operations a planner needs, namely the dimension and a random point, the exponential and logarithmic maps and the geodesic between two configurations, and the metric inner product, norm, and distance.
These operations separate into a base topology and three interchangeable policies (Figure~\ref{fig:infographic}).
The metric supplies \code{inner} and \code{norm}, the retraction supplies the steering map, and the sampler supplies configurations, so any one policy can be replaced without touching the others.
A concrete manifold is a class templated on these policies.
\code{ConfigurationSpace<Base, Metric>} makes the composition explicit by taking its topology from a base manifold and its geometry from an injected metric, without ever calling the base's own metric.
For example, a Euclidean topology paired with a kinetic-energy metric is the configuration space of an articulated arm under joint limits whose distance is measured in kinetic energy rather than in joint angle (Figure~\ref{fig:compose}a).
Table~\ref{tab:manifolds} lists all the topologies, metrics, and retractions \geodex ships.

\looseness=-1
Because the planner is generic over the concept rather than any particular manifold, instantiating it on a concrete type resolves every geometric call at compile time, so the compiler inlines each one into the planner's inner loop instead of dispatching it through a virtual call.
This matters because a single planning query performs hundreds of thousands to millions of distance evaluations, and static dispatch with inlining removes the per-call overhead so that the metric evaluation itself, not the abstraction, dominates the cost.
Under an injected metric, the exact geodesic distance would require solving a boundary-value problem on every call.
\geodex instead approximates it with the metric length of the connecting tangent at the geodesic midpoint, following \cite{kyaw2026geometry}.
Note that this approximation is locally third-order accurate and exact only when the exponential and logarithmic maps are exact under the given metric.
The same types are also available from Python, whose bindings mirror the \mbox{C++} concept through nanobind~\cite{jakob2022nanobind}, so, for example, a configuration-dependent metric is a single callable subroutine that reaches the compiled core through a type-erased wrapper satisfying the \code{RiemannianManifold} concept, letting a planning query be prototyped in a few lines of code while the geometry still runs in \mbox{C++} (Fig.~\ref{fig:compose}b).

\begin{table}[t]
\centering
\caption{
Manifolds and metrics provided by \geodex with the retractions available for each. Every manifold models the \code{RiemannianManifold} concept and pairs with any compatible metric and sampler. A dagger ($\dagger$) marks a cheaper first-order retraction; all others are exact under the default Riemannian metric, and products compose block-wise with exact product distance.
A configuration-dependent metric is a single user callback, and a weighted sum of metrics is again a metric.
}
\label{tab:manifolds}
\footnotesize
\setlength{\tabcolsep}{4pt}
\begin{tabular}{@{}llll@{}}
\toprule
Manifold & Dim & Topology & Retractions \\
\midrule
Euclidean $\Real^n$   & $n$    & flat, optional bounds        & addition \\
Torus $\mathbb{T}^n$           & $n$    & periodic, wrapped            & wrapped addition \\
Sphere $\mathbb{S}^n$          & $n$    & round, ambient $\Real^{n+1}$ & exp, projection$^\dagger$ \\
$\LieGroupSO{2}$      & $1$    & circle group $\cong \mathbb{S}^1$     & group exp \\
$\LieGroupSO{3}$      & $3$    & rotation group               & left/right exp \\
$\LieGroupSE{2}$      & $3$    & planar rigid group           & left/right exp, Euler$^\dagger$ \\
$\LieGroupSE{3}$      & $6$    & spatial rigid group          & left/right screw exp \\
Product $\prod \M_i$  & sum    & direct product               & block-wise \\
ConfigurationSpace    & varies & base $+$ injected metric     & inherited \\
\midrule\midrule
Metric & \multicolumn{3}{l}{Definition} \\
\midrule
Identity              & \multicolumn{3}{l}{flat or round, zero storage} \\
Weighted              & \multicolumn{3}{l}{a configuration-dependent scaling} \\
Kinetic energy        & \multicolumn{3}{l}{$\Transpose{\dot{q}}\MassMatrix(q)\dot{q}$, the mass matrix} \\
Clearance             & \multicolumn{3}{l}{from a signed distance field} \\
$\LieGroupSE{n}$ invariant & \multicolumn{3}{l}{left-invariant, car-like} \\
Affine combined       & \multicolumn{3}{l}{a weighted sum of the above} \\
\bottomrule
\end{tabular}
\vspace{-4mm}
\end{table}

\subsection{Sampling on manifolds}
\label{subsec:sampling}

\begin{figure}[b]
\vspace{-3mm}
\centering
\begin{minipage}{0.3\linewidth}
  \centering
  \includegraphics[width=\linewidth]{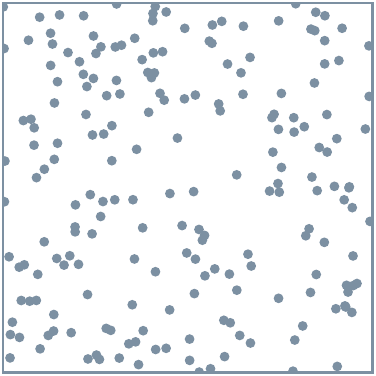}
\end{minipage}\hfill
\begin{minipage}{0.3\linewidth}
  \centering
  \includegraphics[width=\linewidth]{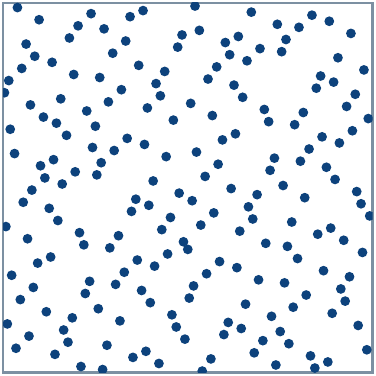}
\end{minipage}\hfill
\begin{minipage}{0.3\linewidth}
  \centering
  \includegraphics[width=\linewidth]{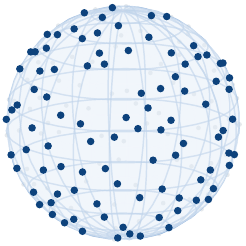}
\end{minipage}
\caption{Low-discrepancy sampling on manifolds. Pseudo-random points on the unit square clump and leave gaps (left), while a Halton sequence of the same size spreads evenly (center). The measure-preserving \code{from\_unit\_cube} map then carries these same points onto the sphere, shown here as the equal-area map from the unit square to $\mathbb{S}^2$ (right), so the points that are uniform in the square remain uniform on the manifold.}
\label{fig:sampling}
\end{figure}

\looseness=-1
\geodex generates samples in two stages.
A sequence generator first produces points in the unit cube $[0,1)^d$, and these points are then mapped to the manifold. Only the second step depends on the geometry of the manifold, so any generator can be used with any manifold. 
Three sequence generators ship with the library: an i.i.d.\ \code{PseudoRandom} baseline, a deterministic \code{Halton} sequence, and a randomized \code{ScrambledHalton} sequence as the default sampler.
The second stage is a measure-preserving map \code{from\_unit\_cube} that every manifold provides.
This map takes the uniform measure on the cube to the intrinsic volume measure of the manifold, so uniform points from a low-discrepancy sequence arrive as uniform points on the manifold (Fig.~\ref{fig:sampling}).

\subsection{Metrics, heuristics, and planners}
\label{subsec:metrics}
\looseness=-1
The metric specifies the geometry of a planning problem, and in \geodex it is a separate, interchangeable object rather than a fixed property of the space.
A configuration-dependent Riemannian metric is given by a single callback that maps a configuration to a symmetric positive-definite matrix.
For example, the kinetic-energy metric takes that matrix from the mass matrix of the robot, so the shortest paths under this metric are minimum-inertia motions.
Any nonnegative combination of metrics is again a metric, so several objectives such as low kinetic energy and high obstacle clearance combine into a single geometry.
Table~\ref{tab:manifolds} summarizes the metrics \geodex provides.

A change of metric also changes whether the heuristics a planner uses remain admissible.
An admissible heuristic requires a lower bound on the cost-to-go, and under a configuration-dependent metric, the usual Euclidean distance is no longer such a bound.
\geodex therefore derives the bound from the metric itself and precomputes it once per metric rather than once per planning query, so the heuristic remains admissible regardless of the geometry a user provides~\cite{kyaw2026loewner}.

\looseness=-1
Rather than reimplementing planners from scratch, \geodex reuses those of OMPL~\cite{sucan2012open} through an adapter that exposes a \geodex manifold as an OMPL state space, so the subroutines the planner uses respect the underlying geometry.
This adapter is the single point at which a call is dispatched at run time;  farther down the stack, every geometric operation is statically dispatched and inlined, as described in Section~\ref{subsec:concepts}.
The same planner source therefore runs on every manifold of Table~\ref{tab:manifolds}, with only the \code{space} argument of \code{plan(space, start, goal)} changing.
\geodex ships with the anytime, asymptotically optimal G-RRT*~\cite{kyaw2024greedy} as its default planner, and a user can switch to a different planner without changing anything else.
Once a solution is found, \geodex post-processes it with randomized shortcutting and L-BFGS smoothing under the same metric.

\subsection{Fast robot dynamics, collision, and integration}
\label{subsec:performance}
\looseness=-1

\geodex includes built-in support for articulated robots through the
kinetic-energy metric, which is given by the mass matrix.
Because this matrix must be evaluated at every configuration visited
during the search, a general-purpose dynamics library is too slow for
real-time planning.
The \code{geodex\_robots} module therefore provides a precompiled evaluator for each supported robot, tracing the mass matrix computation of Pinocchio~\cite{carpentier2019pinocchio} symbolically with CppAD and compiling the result into code that does not require any dynamics library at run time.
Each robot also comes with a lower bound on its mass matrix, computed once at build time and valid over its joint limits, so the heuristic of Section~\ref{subsec:metrics} evaluates from a constant matrix rather than from the mass matrix during the search.
\geodex currently supports the Panda, UR5, Fetch, Baxter, and PR2.

\looseness=-1
Collision checking is specialized in the same way, and the \code{geodex\_vamp} module wraps VAMP~\cite{thomason2024motions} for SIMD-vectorized checking of the same five robots.
Adding a new robot is straightforward, since the evaluator, its bound, and the collision-checking kernels are all generated from a URDF at build time.
\geodex supports Python out of the box, covering the same manifolds, metrics, and planning queries, so a user can prototype a new geometry without writing any \mbox{C++}.
It also provides integrations with the standard ROS~2 stacks, through a Nav2~\cite{macenski2020nav2} global-planner plugin that plans directly on the $\LieGroupSE{2}$ manifold and a MoveIt~2 plugin for manipulators, so metric planning reaches real robotic platforms.
\geodex is open source, tested in continuous integration, and well documented, and can be installed in a single command with pixi or from PyPI.

%% file: sections/evaluation.tex
\section{Evaluation}
\label{sec:eval}

\subsection{Shelf manipulation with a Franka manipulator}
\label{subsec:manipulation}

\begin{figure}[t]
\centering
\includegraphics[width=0.9\linewidth]{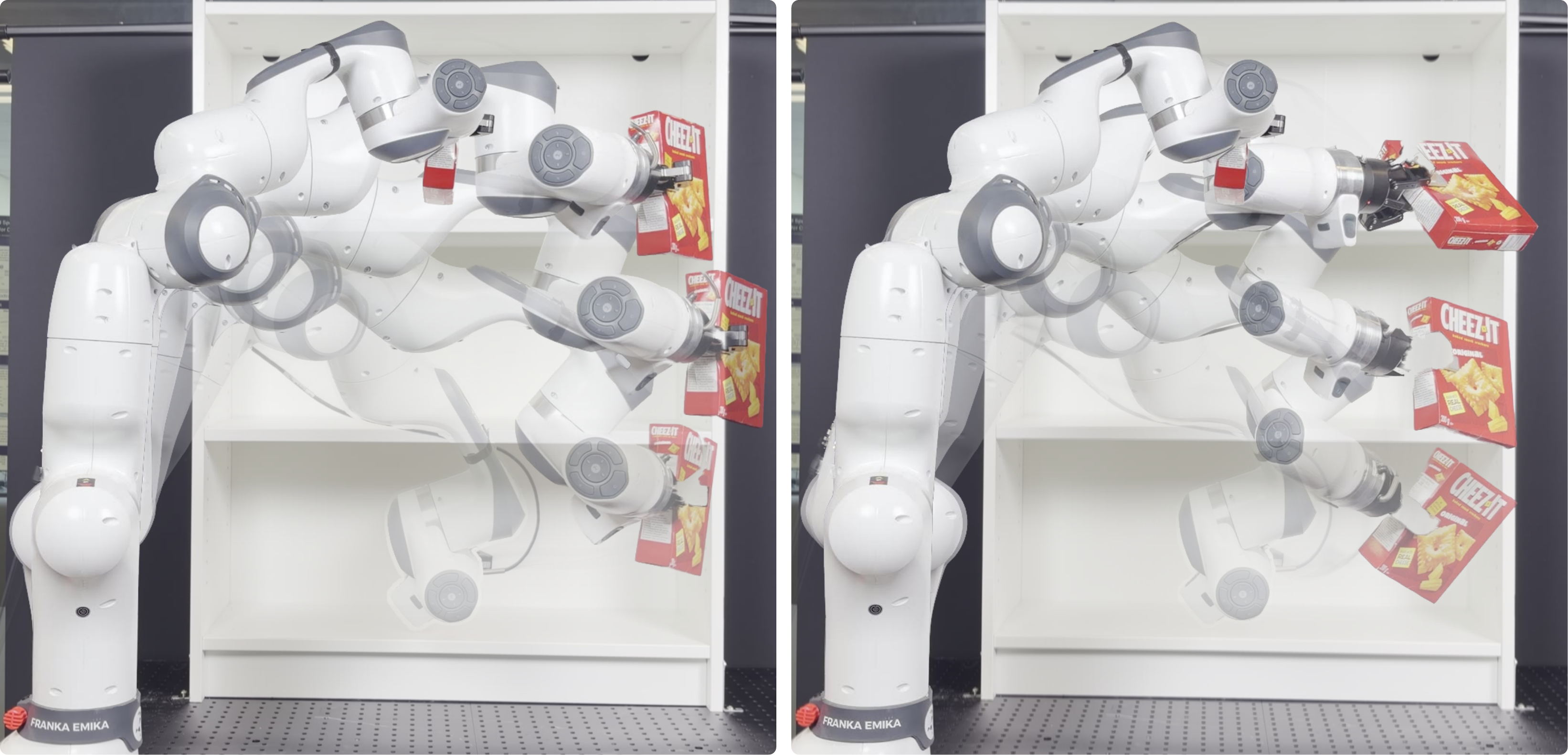}
\caption{
Shelf manipulation with a Franka robot.
The arm moves a cracker box from the bottom compartment to the top one under the Euclidean metric (left) and under the kinetic-energy metric (right), with configurations overlaid along each execution.
The two runs differ only in the metric; with the kinetic-energy metric, the arm turns its heavier joints less and its lighter joints more.}
\label{fig:shelf}
\vspace{-4mm}
\end{figure}

\looseness-1
We first apply \geodex to a downstream manipulation task. We perform shelving with a Franka arm using the MoveIt~2 plugin of \geodex, and evaluate plans using two different metrics without touching the underlying planner or the rest of the planning stack.
Under the Euclidean metric, all directions of motion have the same cost,
so the planner returns the shortest motion in joint space.
Instead, we could also measure distance differently and plan the same task under a different metric. For example, using the kinetic-energy metric, based on the mass matrix of the arm, the shortest paths become minimum-inertia motions~\cite{kyaw2026geometry}.
The heavier joints then rotate only slightly, and the lighter joints farther along the chain rotate significantly more.

\looseness=-1
We compare against three existing motion planners from popular planning frameworks.
The default MoveIt2~\cite{coleman2014moveit} pipeline plans with RRT-Connect and post-processes the result with randomized shortcutting and B-spline smoothing, which reduces motion in joint space under the canonical Euclidean metric.
CHOMP~\cite{zucker2013chomp}, also shipped with MoveIt2, optimizes a joint-space smoothness cost together with an obstacle penalty derived from a signed distance field.
cuRoboV2~\cite{sundaralingam2026curobov2} optimizes a batch of trajectories on a GPU under joint limits, collisions, and a mechanical-work cost built from inverse dynamics.
We run our own plugin twice, once under the Euclidean metric and once under the kinetic-energy metric, and the two runs differ only in the metric handed to the plugin.

We evaluate all planners on a 7-DoF Franka manipulator mounted in front of a shelf with three compartments, bottom (B), middle (M), and top (T).
The task is to move a cracker box from one compartment to another, which gives six problems in total (Figure~\ref{fig:shelf}).
For each problem and planner, we plan and execute the motion five times on the real arm, and take the median of each quantity.
We reparameterize each executed trajectory to a common duration at constant speed, and report the energy the arm spends as its kinetic energy.
We also measure the length of every executed trajectory under both metrics, since the planners minimize different objectives.

Table~\ref{tab:shelf} lists the results, and Figure~\ref{fig:shelf}
shows the box moving from the bottom compartment to the top compartment
in one trial under each metric.
CHOMP is initialized with the RRT-Connect solution, because it cannot
solve these problems from its default straight-line initialization.
The MoveIt~2 implementations of RRT-Connect and CHOMP still fail in
some trials.
cuRoboV2 and \geodex, in contrast, solve every problem, and cuRoboV2 reaches a near-optimal path in the shortest planning time, since it optimizes many initial guesses at once on a GPU.
\geodex finds initial feasible solutions in a few milliseconds, even faster than the GPU-based planner, though its solutions are longer than the ones cuRoboV2 returns.
Since the default planner of \geodex is the anytime G-RRT*~\cite{kyaw2024greedy}, it closes that gap and reaches a comparable or shorter path as we give it more time.

Importantly, using a different metric in \geodex  does not require any changes to the planner itself.
Under the kinetic-energy metric, \geodex returns shorter paths than any other planner we tested, according to that metric.
In the experiments on the real Franka manipulator, the motions from
\geodex also have the lowest accumulated kinetic energy on all six
problems, over five trials each.
The \geodex plugin for MoveIt~2 therefore produces lower-energy motions
than the pipelines included with MoveIt~2.

\begin{table}[t]
\centering
\caption{
Shelf manipulation over the six problems, five trials each.
Success is the fraction of runs that return a valid path.
Time is the planning time for the baselines and the budget for our anytime sampling-based planner.
We measure length under both metrics, after shortcutting and smoothing.
Energy is the median kinetic energy of the executed motion on the real robot, and Best counts how many of the six problems a planner solves with the least energy.
cuRoboV2 runs on a GPU and the others on a single CPU core.
}
\label{tab:shelf}
\scriptsize
\setlength{\tabcolsep}{3pt}
\renewcommand{\arraystretch}{1.12}
\begin{tabular}{lrrrrrr}
\toprule
& & & \multicolumn{2}{c}{\makebox[0pt]{Length$\,\downarrow$}} & \multicolumn{2}{c}{\makebox[0pt]{Energy$\,\downarrow$}} \\
\cmidrule(lr){4-5}\cmidrule(lr){6-7}
Planner & Succ.$\,\uparrow$ & Time & \multicolumn{1}{c}{Euclid.} & \multicolumn{1}{c}{Kinetic} & \multicolumn{1}{c}{Median} & \multicolumn{1}{c}{Best} \\
\midrule\midrule
MoveIt2 RRT-Connect & 80.0\% & 351\,ms & 3.521 & 3.174 & 1.503 & 0/6 \\
MoveIt2 CHOMP & 63.3\% & 116\,ms & 3.554 & 3.804 & 2.768 & 0/6 \\
cuRoboV2 (GPU) & \textbf{100.0\%} & 185\,ms & 1.611 & 1.841 & 0.661 & 0/6 \\
\midrule
\rowcolor{gray!12}
\multicolumn{7}{l}{\textit{\geodex, Euclidean metric (Ours)}} \\
\rowcolor{cFillBlue}
initial & \textbf{100.0\%} & 54\,ms & 4.656 & 3.434 & -- & -- \\
\rowcolor{cFillBlue}
anytime & \textbf{100.0\%} & 250\,ms & 1.712 & 1.880 & -- & -- \\
\rowcolor{cFillBlue}
anytime & \textbf{100.0\%} & 1\,s & \textbf{1.604} & 1.819 & 0.671 & 0/6 \\
\midrule
\rowcolor{gray!12}
\multicolumn{7}{l}{\textit{\geodex, kinetic-energy metric (Ours)}} \\
\rowcolor{cFillBlue}
initial & \textbf{100.0\%} & 88\,ms & 4.418 & 2.623 & -- & -- \\
\rowcolor{cFillBlue}
anytime & \textbf{100.0\%} & 250\,ms & 5.205 & 1.515 & -- & -- \\
\rowcolor{cFillBlue}
anytime & \textbf{100.0\%} & 1\,s & 5.01 & \textbf{1.412} & \textbf{0.507} & \textbf{6/6} \\
\bottomrule
\end{tabular}
\vspace{-4mm}
\end{table}

\subsection{Nonholonomic navigation on $\LieGroupSE{2}$}
\label{subsec:navigation}

\begin{figure}[b]
\vspace{-3mm}
\centering
\includegraphics[width=0.7\linewidth]{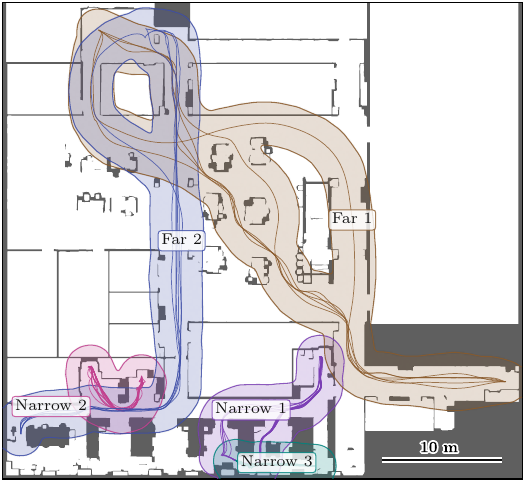}
\caption{Map (roughly $1,600~\mathrm{m}^2$) used for nonholonomic navigation on $\LieGroupSE{2}$. White areas are traversable, and gray indicate obstacles. We also illustrate areas traversed by the robot in the real-robot experiments (Sec.~\ref{sec:se2_real_robot}). The \emph{narrow} areas are inside a cluttered office space, and the \emph{far} areas a publicly accessible atrium space in a university building.}
\label{fig:nav-map}
\end{figure}

\looseness=-1
Our second use case applies \geodex to mobile robot navigation on the Lie group $\LieGroupSE{2}$.
We compare against three standard graph-based planners from the ROS~2 navigation stack, Nav2~\cite{macenski2020nav2}.
Theta*~\cite{daniel2010theta} searches for any-angle paths on the costmap grid in $\Real^2$ without considering the heading of the robot; we additionally smooth its output as a post-processing step.
The other two baselines are built on discretized motion primitives~\cite{macenski2024open}.
The Smac state-lattice planner searches a precomputed control set~\cite{pivtoraiko2009differentially}, while the Smac hybrid-A* planner expands Dubins and Reeds-Shepp primitives from a grid of discretized headings~\cite{dolgov2010path}.
We evaluate all four planners on a campus building map roughly $40m\times40m$ (Figure~\ref{fig:nav-map}), which we build at both 2\,cm and 5\,cm resolution using a Clearpath Jackal \cite{samavi2025deploying} running Cartographer \cite{hess2016cartographer}.
The building has both long open areas and cluttered office spaces, and we run one scenario for each.

\subsubsection{Offline experiments}
\looseness=-1
In the \emph{far} scenario (e.g., brown and blue in Figure~\ref{fig:nav-map}), the start and the goal poses lie across the whole map, so we can compare how fast planners find solutions over long routes.
In the \emph{narrow} scenario, both poses sit in the aisles of a cluttered office area (e.g., purple, magenta and green in Figure~\ref{fig:nav-map}), so we can measure how reliably a planner finds a path through a gap barely wider than the robot.
We generate 25 start and goal pairs at random in each scenario, and we perturb each pose into eight variants by up to 0.04~m in position and 45 degrees in heading to evaluate how sensitive a planner is to the exact start and goal.
This gives 400 distinct queries, which we run at both resolutions, ten times for every planner, with different pseudorandom seeds for our stochastic planner, resulting in 32,000 runs in total.
We report the success rate, the collision rate, the planning time, the path length in $\Real^2$, the length of the same path under the $\LieGroupSE{2}$ metric, and the smoothness.

\begin{table*}[t]
\centering
\caption{
Planning results on the building map for the far (left) and narrow (right) scenarios, with each cell giving 2\,cm\,/\,5\,cm resolution. Succ.: \% of queries returning a path. Coll.: \% of those paths in collision. Other columns are medians over queries solved by all planners, with time in ms, $\Real^2$ length in m, $\LieGroupSE{2}$ length under the plugin metric, and smoothness in rad/m.
}
\label{tab:nav}
\scriptsize
\setlength{\tabcolsep}{3pt}
\newcolumntype{P}{r<{/\hspace{-\tabcolsep}}>{\hspace{-\tabcolsep}}l}
\begin{tabular}{@{}l *{6}{P} >{\hspace{4pt}}P *{5}{P}@{}}
\toprule
& \multicolumn{12}{c}{\textit{Far scenarios}} & \multicolumn{12}{c}{\textit{Narrow scenarios}} \\
\cmidrule(lr){2-13}\cmidrule(l){14-25}
Planner
& \multicolumn{2}{c}{Succ.$\,\uparrow$} & \multicolumn{2}{c}{Coll.$\,\downarrow$} & \multicolumn{2}{c}{Time$\,\downarrow$} & \multicolumn{2}{c}{Len.\,$\Real^2\,\downarrow$} & \multicolumn{2}{c}{Len.\,$\LieGroupSE{2}\,\downarrow$} & \multicolumn{2}{c}{Smooth.$\,\downarrow$}
& \multicolumn{2}{c}{Succ.$\,\uparrow$} & \multicolumn{2}{c}{Coll.$\,\downarrow$} & \multicolumn{2}{c}{Time$\,\downarrow$} & \multicolumn{2}{c}{Len.\,$\Real^2\,\downarrow$} & \multicolumn{2}{c}{Len.\,$\LieGroupSE{2}\,\downarrow$} & \multicolumn{2}{c}{Smooth.$\,\downarrow$} \\
\midrule
Theta*
& \textbf{100.00} & \textbf{100.00} & 0.00 & 0.00 & 363 & 87 & \textbf{38.75} & \textbf{38.82} & 54.37 & 48.15 & 0.602 & 0.381
& 100.00 & 100.00 & \textcolor{cGeo}{30.67} & \textcolor{cGeo}{27.67} & \textbf{24} & 9 & \textbf{6.52} & \textbf{6.48} & 15.99 & 14.40 & 1.453 & 1.151 \\
Smac-Lattice
& 88.48 & 88.48 & \textcolor{cGeo}{0.50} & 0.00 & 179 & 55 & 40.02 & 39.32 & 51.49 & 51.66 & 0.319 & 0.308
& 73.00 & 73.50 & 0.00 & \textcolor{cGeo}{1.50} & 34 & 15 & 6.95 & 7.06 & 14.94 & 14.50 & 0.834 & 0.848 \\
Smac-HybA*
& \textbf{100.00} & \textbf{100.00} & 0.00 & 0.00 & 101 & \textbf{26} & 39.45 & 39.30 & 50.60 & 49.02 & 0.444 & 0.396
& 94.50 & 94.50 & \textcolor{cGeo}{1.00} & \textcolor{cGeo}{2.00} & 32 & \textbf{9} & 7.03 & 6.79 & 13.73 & 13.26 & 0.837 & 0.829 \\
\rowcolor{cFillBlue}
\textbf{\geodex (Ours)}
& \textbf{100.00} & \textbf{100.00} & 0.00 & 0.00 & \textbf{72} & 48 & 41.79 & 41.82 & \textbf{47.00} & \textbf{47.15} & \textbf{0.219} & \textbf{0.223}
& \textbf{100.00} & \textbf{100.00} & 0.00 & 0.00 & 53 & 37 & 7.12 & 7.18 & \textbf{11.64} & \textbf{11.78} & \textbf{0.612} & \textbf{0.614} \\
\bottomrule
\end{tabular}
\end{table*}
\begin{figure*}[t]
\centering
\begin{minipage}{0.33\linewidth}
  \centering
  \includegraphics[width=\linewidth]{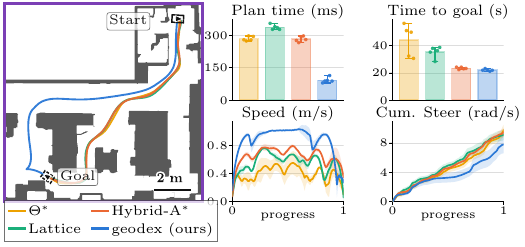}
  \vspace{-6mm}
  \subcaption{Narrow 1: doorway, reverse into the goal.}
  \label{fig:narrow1}
\end{minipage}\hfill
\begin{minipage}{0.33\linewidth}
  \centering
  \includegraphics[width=\linewidth]{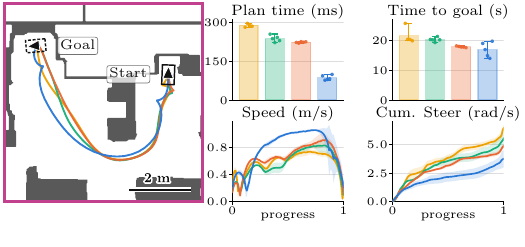}
  \vspace{-6mm}
  \subcaption{Narrow 2: tight start pocket, U-turn.}
  \label{fig:narrow2}
\end{minipage}\hfill
\begin{minipage}{0.33\linewidth}
  \centering
  \includegraphics[width=\linewidth]{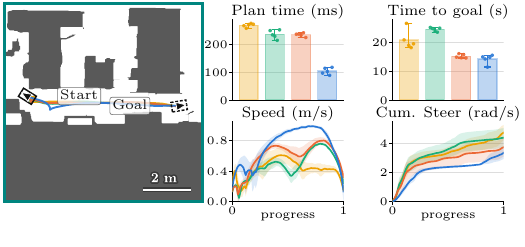}
  \vspace{-6mm}
  \subcaption{Narrow 3: corridor, back out of the start.}
  \label{fig:narrow3}
\end{minipage}
\vspace{-1mm}
\caption{Narrow scenarios on the Jackal. Left of each panel: the global costmap at the time of planning, the start (solid) and goal (dashed) footprints, and one exemplary executed trajectory per planner. Right: planning time and time to goal (bars are the mean, whiskers span the full range, dots are individual runs), and the mean with one standard deviation of speed and cumulative steering effort $\int|\dot\omega|\,dt$ against normalized progress along the executed path. \geodex plans $2.5$ to $3\times$ faster, arrives first, holds a higher cruise speed, and accumulates less steering effort, because it optimizes on $\LieGroupSE{2}$ rather than a Euclidean surrogate.}
\label{fig:narrow_nav}
\vspace{-3mm}
\end{figure*}
\begin{figure}[t]
\centering
\begin{minipage}{0.9\linewidth}
  \centering
  \includegraphics[width=\linewidth]{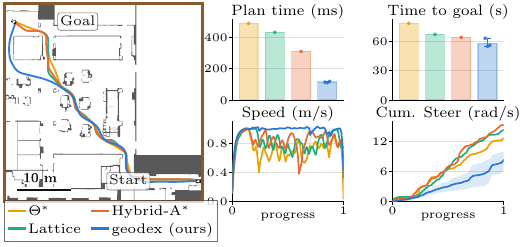}
  \vspace{-6mm}
  \subcaption{Far 1: \SI{36}{\meter} across the floor, open office to corridor.}
  \label{fig:far1}
\end{minipage}
\vspace*{1mm}

\begin{minipage}{0.9\linewidth}
  \centering
  \includegraphics[width=\linewidth]{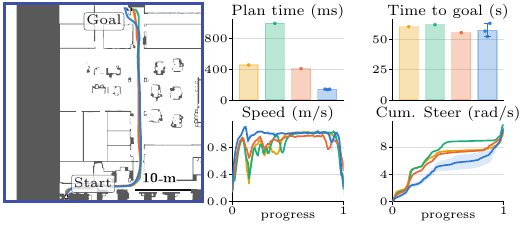}
  \vspace{-10mm}
  \subcaption{Far 2: \SI{36}{\meter} along the main corridor and around the core.}
  \label{fig:far2}
\end{minipage}
\vspace{-1mm}
\caption{Far scenarios on the Jackal. Over these routes, the planning-time gap between \geodex and the baselines widens to $3$ to $7\times$, and the baselines shed speed at every corner while \geodex holds its cruise speed, accumulating roughly half the steering effort. The baselines were run once per scenario.}
\label{fig:far_nav}
\vspace{-7mm}
\end{figure}

Table~\ref{tab:nav} summarizes the results.
The two Smac planners are the only ones that fail to return a path, and they fail most often in the narrow scenario.
Both expand a pose only through precomputed motion primitives, so a small change in the position or heading of a planning query changes the set of expansions available to them.
They also cannot reach the goal exactly, since both require a large goal tolerance, set by the resolution of the discretization, before they return a path at all.
Theta* solves every query and returns the shortest paths in $\Real^2$, but since it searches only the plane rather than $\LieGroupSE{2}$, its paths end up in collision, especially in the narrow scenario.
Since Theta* expands a fixed-size grid, its planning time also grows with the grid resolution and with the distance between the start and the goal.
Our planner, on the other hand, solves every query without collision, and the grid resolution does not affect its planning time, since we use the grid for collision checking only.
It reaches the goal exactly, since it plans directly on the manifold the robot operates on.
\geodex also returns the smoothest paths of all four planners, and it works reasonably well in the tight spaces where existing planners built on motion primitives are known to fail.

\subsubsection{Real-world experiments} \label{sec:se2_real_robot}
We use the same map using the Jackal's autonomy stack and compute hardware based on \cite{samavi2025deploying, samavi2025sicnavdiffusion}. To track each plan, we implement a model predictive controller (MPC) replanning at 10 Hz and optimizing a 2s horizon along the plan. At each MPC replan step, we first extract a 2s reference from the plan based on the current pose of the robot by selecting the starting point of the reference to be the closest point on the $\LieGroupSE{2}$ plan via the left-invariant metric. We then build the rest of the reference by interpolating along the plan at the closest speed to a specified target (1.2 m/s) admitted by the robot's kinodynamic constraints (acceleration limit, yaw-rate limit, max. lateral robot wheel slip). The MPC cost penalizes error versus the reference using quadratic longitudinal, lateral, and heading cost terms and penalizes controller effort using an input rate penalty. In order to ensure collision avoidance, we also add a discrete-time control barrier function on the signed distance field of the live 2cm local costmap \cite{dorazio2026sm2ith}. We use this controller setup with the same parameters for all evaluated planners.

We evaluate three narrow scenario start-and-end-goal pose pairs (highlighted in purple, magenta and green in Figure~\ref{fig:nav-map}) and repeat five runs using each planner, resulting in 60 total runs. We also evaluate two far scenarios (highlighted brown and blue in Figure~\ref{fig:nav-map}) and conduct one run for each baseline and three runs for \geodex, resulting in 12 total runs. 
To evaluate robot efficiency, we measure the time it takes each planner to find and optimize a path (plan time) and the time it takes for the robot to track the path. To evaluate task execution quality, we measure the linear speed of the robot and the cumulative steering effort along each path.

Figure~\ref{fig:narrow_nav} summarizes the results of three narrow scenarios in an office environment. We plot the costmap with an example plan for each planner on the left, bar plots for the efficiency metrics on the top right and the mean with one standard deviation of robot linear speed and cumulative steering effort over the path on the bottom right. In all three scenarios, \geodex consistently outperforms the baselines in terms of total planning time, critically, remaining below 100 ms, allowing real-time replanning in field robotics scenarios. \geodex is also fastest to the goal in all three scenarios, although the margin over Smac Hybrid-A* is small.
In terms of execution quality, \geodex operates closer to the target linear speed, holding 60\% to 72\% of 1.2 m/s while the baselines hold only 33\% to 54\%. Furthermore, we can observe that following the \geodex plan results in a smoother speed profile with fewer high-frequency speed changes throughout the path as the robot controller does not need to constantly slow down to steer. The cumulative steering effort of the \geodex runs is lower in all three scenarios, and the curves separate gradually over the whole path rather than at one point, indicating that the difference is due to the shape of the plan rather than a single maneuver.

\looseness=-1
Figure~\ref{fig:far_nav} reports two longer routes of roughly 36 m across. Compared to the narrow scenarios, we observe that the planning-time gap widens between our method and baselines to 115 ms and 139 ms against 310 ms to 990 ms. Over the long distances, the speed profiles separate most clearly, with the baselines dipping at every corner while the robot following the \geodex plan holds its target speed throughout, and cumulative steering effort in the first scenario is roughly half that of the baselines.
Time to goal follows the same pattern as the narrow scenarios in Far 1 (58.0 s against 64.2 s to 78.4 s) but is within noise in the second, where the route is a single long corridor that offers little opportunity for our more smoothly-shaped plans to pay off.

%% file: sections/conclusion.tex
\section{Conclusion}
\label{sec:conclusion}

\looseness=-1
We presented \geodex, an open-source library that treats the configuration-dependent Riemannian metric as the geometry that drives planning.
A user picks a manifold and a metric, and the same planner runs on that space without requiring any change to its search.
Distance functions, interpolation, and the heuristics that planning relies on all follow from the metric, so changing to a new geometry only needs a new metric rather than rewriting a planner.
We demonstrated the capabilities of \geodex on two downstream real robot use cases.
On a Franka manipulator, swapping the Euclidean metric for the kinetic-energy metric changes how the arm moves a box between shelf compartments, while the planner and the rest of the stack stay untouched.
On a differential-drive robot, weighting sideways motion in the metric replaces the use of motion primitives that existing planners expand, and the resulting paths are smoother and faster for the robot to follow.
We would argue that, although we demonstrate \geodex here for nonholonomic planning, nothing stops us from using it for any motion model by simply adjusting the metric to mimic holonomic or Ackermann models, while keeping the smoothness and planning-time benefits of the manifold-based approach.

\geodex plans only on manifolds that a user can parameterize, so it cannot yet handle task constraints such as holding an end effector on a surface.
We plan to support constraint manifolds next, so that the same metric-first interface extends to planning under end-effector task constraints.
Our planner also evaluates the metric on the CPU, one query at a time, and we are interested in moving these evaluations onto the GPU, where many of them can run at once.
We make \geodex publicly available, together with the MoveIt2 and Nav2 plugins and all the benchmarks used in this paper.